# Task Embedding Temporal Convolution Networks for Transfer Learning Problems in Renewable Power Time-Series Forecast

Jens Schreiber[1][0000−0002−9979−8053], Stephan Vogt[1], and
Bernhard Sick[1][0000−0001−9467−656X]

University of Kassel, Wilhelmshöher Allee 71, 34121 Kassel, Germany
{j.schreiber, stephan.vogt, bsick}@uni-kassel.de

**Abstract.** Task embeddings in multi-layer perceptrons for multi-task learning and inductive transfer learning in renewable power forecasts have recently been introduced. In many cases, this approach improves the forecast error and reduces the required training data. However, it does not take the seasonal influences in power forecasts within a day into account, i.e., the diurnal cycle. Therefore, we extended this idea to temporal convolutional networks to consider those seasonalities. We propose transforming the embedding space, which contains the latent similarities between tasks, through convolution and providing these results to the network's residual block. The proposed architecture significantly improves up to 25% for multi-task learning for power forecasts on the EuropeWindFarm and GermanSolarFarm dataset compared to the multi-layer perceptron approach. Based on the same data, we achieve a ten percent improvement for the wind datasets and more than 20% in most cases for the solar dataset for inductive transfer learning without catastrophic forgetting. Finally, we are the first proposing zero-shot learning for renewable power forecasts to provide predictions even if no training data is available.

## 1 Introduction

The Paris commitment demands to limit human-induced global warming below 2°C above pre-industrial levels to reduce the impact of the climate crisis. To achieve the commitment, renewable energy resources need to increase their share from 14% in 2015 to 63% in 2050 in the worldwide energy production [1]. One central problem with renewables, such as wind and photovoltaic (PV) parks, is that they are not controllable like traditional energy supplies. Instead, they are weather dependent and storage capabilities for unexpected power generation are limited. Therefore, to assure grid stability, energy suppliers require reliable power forecasts based on numerical weather predictions (NWPs). These forecasted weather features, such as wind speed or radiation, are the input to models predicting the expected power generation in *day-ahead* forecasts between 24 and 48h into the future.

Another problem is that each of those parks typically has individual characteristics to learn by the forecast models. Therefore, a single forecast model is learned per park or even for a single wind turbine in practice. Assuming there are about 30,000 wind facilities with an average number of 10,000 parameters per model solely in Germany, this makes a total of 300 million parameters to train. Additionally, the hyperparameters need to be optimized, e.g., through grid-search.

The training of those models is in some sense contradictory to the Paris commitment, as the training and even the inference themselves have an extensive energy demand and cause a considerable amount of carbon emmission [2]. To reduce the number of models that need to be optimized while leveraging knowledge between parks, [3] proposed to utilize multi-task learning (MTL) approaches. In their task embedding multi-layer perceptron (MLP), the discrete task information, *which task is to be predicted*, is encoded through an embedding layer and concatenated with other input features. However, this approach still requires a reasonable amount of training data. This training data is often not available for new parks. Therefore, in [4] an inductive transfer learning (TL) experiment is evaluated. In this inductive transfer learning setting, the task embedding MLP extracts knowledge from a set of source tasks. This knowledge is adapted to unseen target tasks, during initial training, with limited training instances. In their approach, a Bayesian variant of the task embedding MLP leads to the best results for wind parks with limited training data.

However, both approaches for MTL and inductive transfer learning are missing two critical points. First, there are seasonal patterns, i.e., the diurnal cycle, within day-ahead forecasts affecting the forecast error [5]. Those are not considered within an MLP, even though the substantial benefits of time series forecasts through convolutional neural networks (CNNs) are known [6]. Second, ideally, we can forecast parks without any historical power measurements. This zero-shot learning paradigm is essential to assure reasonable forecasts from the beginning of a new park. Therefore, we are interested in answering the following research questions:

*Question 1.* Can a MTL CNN architecture improve the forecast error for wind and PV parks compared to a similar MLP architecture?

*Question 2.* Are CNN based MTL architectures capable of providing sufficiently good forecasts in zero-shot learning for renewable power forecasts?

*Question 3.* Are CNN based MTL architectures beneficial during inductive TL compared to a similar MLP approach?

Providing an architecture dealing with those questions allows us to tackle practical, real-world challenges during the expansion of renewables. Therefore, we propose task-*temporal convolution network (TCN)*, see Fig. 1. *Task-TCN* encodes task-specific information through an embedding layer. Thus, each park gets an

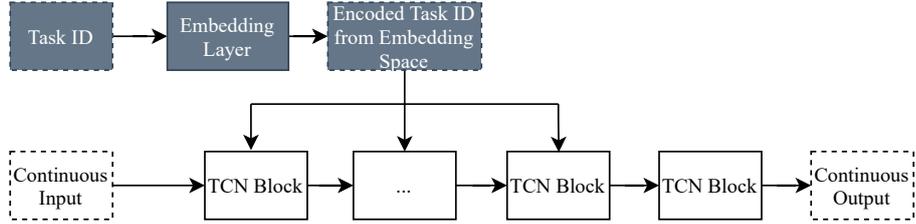

**Fig. 1.** Task-TCN encodes a task ID, for each task, through an embedding layer. The learned encoding from the embedding space is *added* in the residual block to provide task-specific forecasts.

*increasing* task ID ($m \in \mathbb{N}^+$) assigned as input to the embedding layer. We assure that the TCN learns relations between tasks during training by adding the encoded task information in a residual block. By encoding the task ID through an embedding layer, we can extend the network to new parks while avoiding catastrophic forgetting of previous parks. All experiments are conducted on the open EuropeWindFarm and GermanSolarFarm day-ahead datasets[1]. Evaluation of the task-TCN on these datasets, in comparison to the Bayesian task embedding MLP as the baseline, leads to the following contributions:

- The proposed task-TCN architecture, for MTL renewable power forecasts, leads to improvements of up to 25 percent.
- The proposed task-TCN leads to improvements of more than ten percent for wind and up to 40 percent for the solar dataset for inductive TL problems.
- We are the first to propose zero-shot learning in renewable power forecasts to provide forecasts without training data.

The source code and supplementary material of the experiments are openly accessible[2]. The remainder of this article is structured as follows. Section 2 gives details on related work. The following Section 3 introduces relevant definitions and details the proposed approach. We detail the datasets, discuss the experiment and most essential findings in Section 4. In the final section, we summarize our work and provide insides for future work.

## 2 Related Work

To answer our research question, we review related work in the field of MTL, (inductive) TL, and zero-shot learning focusing on power forecasts for wind and PV. Formal definitions of those topics are given in Sec. 3.1.

---

[1] www.uni-kassel.de/eecs/de/fachgebiete/ies/downloads.html, accessed 2021-03-39
[2] github.com/anonymous/anonymous, accessed 2020-03-29

## 2.1  Multi-Task Learning for Renewable Power Forecasts

The authors of [3] propose a task embedding for efficient encoding of task-specific information in a hard parameter sharing (HPS) network improving the forecast error with a minimal amount of parameters for day-ahead wind and solar power forecasts. In an investigation of [7], a self-attention based MTL encoder-decoder architecture is proposed for ultra-short term PV power forecasts. The study of [8] proposes an iterative MTL strategy for Gaussian processes to forecast PV outputs. Their approach allows to impute missing values, extract local irregular components as well as global trends.

## 2.2  Transfer Learning for Renewable Power Forecasts

The authors of [9] train autoencoders for different months of the training data, through TL based on an initially trained model. Forecasts of those autoencoders, along with the original features, are used to train a deep belief network as a meta learner for short-term wind power forecasts. The concept of [10] is an innovative approach for quantile regression based on gradient boosting decision trees. By combining instance based TL with a weighting strategy for source tasks, they achieve substantial improvements upon the baseline. Moreover, the article [11] provides a strategy for quantile regression for day-ahead solar power forecasts. First, an ensemble for day-ahead PV point forecasts is trained through a meta-learning approach. Afterward, the model is finetuned for probabilistic forecasts. To reduce computational costs for wind power forecasts [12] provides a data-driven strategy. Initially, learning a unified autoencoder for all source tasks allows deriving specific features for parks with minimal computation costs. Finally, the models are adopted for the target. Using a recurrent network and finetuning, the authors of [13] achieve good results for multi-step prediction for ultra-short-term forecast horizon of PV. Through a cluster-based domain adaption approach, substantial improvements for TL are achieved in wind power forecasts in [14]. By clustering similar wind parks through their distribution, an intelligent weighting scheme provides predictions for a new park. The approach of [15] uses various machine learning models to predict short-term wind power time series. To neglect the limited data in the target domain, they build an auxiliary dataset through k-nearest neighbors. All algorithms improve through the auxiliary dataset. The authors of [16] use a combination of autoencoders, neural networks, and transfer learning to reduce the training time for short-term predictions. Initially, all autoencoders are trained based on a single wind park and only finetuned for the others. The evaluation results show improved results compared to similar techniques. A thorough evaluation of inductive TL for day-ahead wind power forecast is given in [4]. Furthermore, a Bayesian task embedding for an MLP is proposed and allows to leverage the knowledge from multiple sources in the transfer learning step. The Bayesian task embedding is superior to models learned from scratch and a traditional HPS. However, they neglect the possibility to extend this approach to CNNs.

The related work shows limited research on TL and MTL for day-ahead power forecasts. Furthermore, none of those approaches provide a framework for wind as well as PV power forecasts. Moreover, there is limited research in addressing TL challenges through recent work in time-series forecasts [17, 18]. Further, to the best of our knowledge, there has been no attempt to provide zero-shot forecasts in the field of renewable energies. Besides, none of those proposals consider a unified framework that takes recent advances in time-series forecast into account, such as TCN, while being applicable to MTL, inductive TL, and zero-shot learning.

## 3 Proposed Method

In the following, we detail essential definitions, relate the task embedding MLP to MTL, and describe our proposed architecture.

### 3.1 Definition of MTL, TL, and Zero-Shot Learning

The following definitions have been introduced in [19–21]. We slightly modified them for a consistent formulation of, e.g., multi-source TL and MTL. Before defining inductive TL, zero-shot learning, and MTL, we review the definition of a single task and domain.

**Definition 4 (Domain).** *A domain is defined by $\mathcal{D} = \{\mathcal{X}, P(X)\}$, where $\mathcal{X}$ is the feature space and $P(X)$ is the marginal distribution with $X = \{\mathbf{x} \mid \mathbf{x}_i \in \mathcal{X}, i = 1, \ldots, N\}$.*

**Definition 5 (Task).** *The task of a domain is defined with $\mathcal{T} = \{\mathcal{Y}, f(\cdot)\}$, where the function is defined by $f : \mathcal{X} \to \mathcal{Y}$. The function $f(\cdot)$ is learned by training instances $\{\mathbf{x}_i, y_i\}$ with $\mathbf{x}_i \in X$ and $y_i \in Y$, where $Y = \{y \mid y_i \in \mathcal{Y}, i = 1, \ldots, N\}$. The function $f(\cdot)$ describes characteristics of the distribution $P(Y \mid X)$. In a Bayesian approach those are samples and the expectation in a frequentist view.*

Note, in these definitions, $\mathbf{x}_i$ are available weather predictions and $y_i$ are historical power measurements of a park. We assume that $X$ and $Y$ are ordered sets for the required time-series forecasts for simplicity.

**Definition 6 (Inductive Transfer Learning).** *Inductive Transfer Learning has the goal to transfer knowledge from source (S) tasks $\{\mathcal{T}_{S_m}\}_{m=1}^{m=M}$ to a target (T) task $\mathcal{T}_T$. Therefore, we use $M \in \mathbb{N}^+$ source domains $\{(\mathcal{D}_{S_m}, \mathcal{T}_{S_m}) \mid m = 1, \ldots, M\}$ and (limited) training instances $\{\mathbf{x}_{T_i}, y_{T_i}\}$ with $\mathbf{x}_{T_i} \in X_T$ and $y_{T_i} \in Y_T$ to learn a function $f_T(\cdot)$.*

In contrast to transductive TL, labeled training data in the target domain is available in inductive TL.

**Definition 7 (Zero-shot learning).** *Zero-shot learning can be interpreted as unsupervised transductive TL [22]. In this setting, meta-information from source and target tasks is used to select an appropriate prediction function $f_{S_m}(\cdot)$ to predict the target task. Importantly, no training instance $\{\mathbf{x}_{T_i}, y_{T_i}\}$ is used to finetune the function.*

In comparison to transductive TL, zero-shot learning is not using a domain adaption approach between the source(s) and the target. This unsupervised approach makes it an even more challenging problem as no assumptions are made of the source and target domain.

**Definition 8 (Multi-Task Learning).** *In MTL approaches, each task is accompanied by a domain $\mathcal{D}_m$ with $i \in \{1, \ldots, N\}$ training instances $(\mathbf{x}_i^m, y_i^m)$, where $\mathbf{x}_i^m \in X^m$, $y_i^m \in Y^m$, $m \in 1, \ldots, M$, and $M \in \mathbb{N}^+$ tasks.*

In contrast to inductive transfer and zero-shot learning, all tasks have a sufficient amount of training data in MTL problems. Furthermore, in MTL we are typically interested in improving all tasks' forecast errors simultaneously.

### 3.2    Proposed Method

Typically, in MTL, there are two approaches to share knowledge and improve the forecasts error: soft parameter sharing (SPS) and HPS. In SPS, each task has, e.g., an individual deep learning model and similarity is enforced through regularization [23]. In HPS deep learning architectures, we typically have mostly common layers that are the same for all tasks along with some task-specific output layers. As SPS requires additional parameters compared to HPS, we focus on the latter.

In the previous section, we argued the function $f(\cdot)$ describes the conditional distribution $P(Y \mid X)$, where $Y$ is a set of observed power generations and $X$ are observations of the numerical weather prediction such as wind speed and radiation. However, this description neglects the possibility to model individual forecasts for different parks in MTL settings. Therefore, we describe how to model this dependency through an embedding layer in an MLP as suggested in [3, 4]. By doing so, we are the *first to provide a mathematical description* between the task embedding and the MTL definition. Afterward, we give details on using a Bayesian embedding layer to assure that similar parks are close to one another in the embedding space as proposed in [4]. Finally, we detail how our approach adopts the idea of task embeddings to TCNs.

**Task-Embedding for MLPs** encodes an increasing (discrete) task ID about *which task $m \in \mathbb{N}^+$ is to be predicted* through an embedding layer and concatenates results of this embedding space with other continuous input features [3, 4]. In the following we connect embedding layers to the MTL definition. Therefore, consider a common function $h(\cdot)$ for all $M$ tasks that approximates:

$$P(y_m \mid X_m, g(m)), \tag{1}$$

where $m$ is the discrete information whose task is to be predicted with $m \in 1, \ldots, M$ and $M \in \mathbb{N}^+$. $g(m)$ is then a transformation into an arbitrary real valued dimension.

Assuming we have such a transformation, the conditional modeling allows us to develop a model, e.g., without task-specific layers. The required information on the task is given through $g(m)$. Therefore, the function $h(\cdot)$ for MTL has the training instances $\{\mathbf{x}_i^m, y_i^m, g(m)\}$. Ideally, we want a mapping $g$ that is beneficial for the MTL problems, e.g., similar tasks are close to another in the embedding space. Respectively, an MTL approach needs to learn this mapping function during training. For such a problem, the authors of [24] propose the following equation of an embedding layer:

$$g(m) = \sum_{\alpha=1}^{M} \mathbf{w}_{\alpha\beta} \boldsymbol{\delta}_{m\alpha} = \mathbf{w}_{m\beta}, \qquad (2)$$

where $\boldsymbol{\delta}_{m\alpha}$ is the Kronecker delta. Therefore, $\boldsymbol{\delta}_{m\alpha}$ is a vector of length $M$ ($M$ tasks), where only the entry with $\alpha = m$ is non-zero and $\mathbf{w}_{\alpha\beta} \in \mathbb{R}^D$ is the learnable vector at this position. Respectively, the function $g$ maps the discrete value $m$ of a task ID through a one-hot encoding to a trainable vector.

This transformation is then concatenated with other (continuous) features. Due to the joint training for multiple tasks, e.g., within one batch, we have various tasks, it is beneficial for the network to learn a mapping where similar tasks have a similar vector $\mathbf{w}_{\alpha\beta}$. This mapping allows utilization of a similar transformation in later layers for similar domains and targets.

**Bayesian Task-Embeddings** are especially interesting in the scenario where limited data for a (single) task is available. If a limited amount of data is available, e.g., in the inductive transfer learning problem, a Bayesian approach allows placing an identical, independent prior on $\mathbf{w}_{m\beta}$. This prior ensures that indistinguishable tasks, due to limited data, are within a similar neighborhood. At the same time, tasks with sufficient data have an encoding that is different from other tasks. We apply *Bayes by backprop* solely on the embedding layer. In this way, we minimize the training effort for the number of additional parameters and benefit from the embedding space's Bayesian approach.

**Task-Embedding for Temporal Convolution Neural Networks** are a way to take advantage of seasonal patterns, e.g., the diurnal cycle, in *day-ahead* forecasts [5] in an MTL architecture. In MTL day-ahead power forecast problems, we forecast the expected power $y_{24}^m, \ldots, y_{48}^m$ for $m \in 1, \ldots, M$ parks. In contrast to intra-day forecasts between 0 and 23 hours into the future, day-ahead forecasts are more challenging as the forecast error increases with an increasing forecast horizon [25]. We have the numerical weather features $\mathbf{x}_{24}^m, \ldots, \mathbf{x}_{48}^m$ from a weather prediction originating from $t_0$ (e.g. originating from 0 o'clock UTC) as input. The fundamental building block of task-TCN is the *TCN*, a CNN architecture developed for time series. In this architecture, dilated convolutions ensure that only the past values are considered. To avoid the vanishing gradient problem, the TCN uses residual blocks in each layer, see Fig. 1. In each residual block, the input is processed through two times the following pattern: dilated convolution, weight norm, ReLU activation, and dropout for regularization, see Fig. 2. An optional convolution to transform the original input to the output shape ensures that the initial input features match those of the residual block's output. Finally, this potentially transformed input is added to the other output.

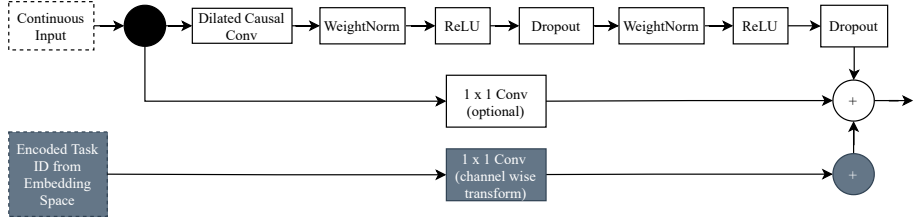

**Fig. 2.** Residual block of the task-TCN. The encoded task ID, from the embedding space, is transformed through a 1D convolution. The transformed task ID is then added to the results of the original TCN residual block.

In the previously discussed task embedding for MLPs, the embedding is concatenated with the continuous weather input features to provide task-specific forecasts. However, simply concatenating the embedding output along the time axis in CNNs leads to additional channels with the same information for each time step. Respectively, having redundant information along the time axis. In preliminary experiments, we found this not to work well due to the redundant information causing a plateau in the training error.

Instead, we propose to *add* the result from $g(m_i)$ at the end of the residual block, see Fig. 2. Therefore, we first transform the encoded task ID through a 1-D convolution layer with a kernel size of one. The transformation gives the network the possibility to learn an encoding that is different in each channel. This principle applies to any residual block within the network. A study on the diverse learned similarities of tasks per channel is presented in the supplementary material. After applying this transformation, the output is *added* to the result of the original residual block.

## 4    Experimental Evaluation of the Task-Temporal Convolution Network

To answer our research questions, we conducted one experiment for each of the three research questions. We evaluate each of these experiments on the datasets explained in Sec. 4.1 along with its pre-processing. Sec. 4.2 lists relevant evaluation measures. The design of experiments and the evaluations are detailed in Sec. 4.3, 4.4, and 4.5. The models detailed in Sec. 4.3 are the source models for the other experiments. The description also includes the Bayesian MLP that is the baseline in all experiments.

### 4.1    GermanSolarFarm and EuropeWindFarm Dataset

In the GermanSolarFarm and the EuropeWindFarm dataset, the uncertainty of the NWP makes it challenging to predict the generated power. We refer to these as wind and solar datasets in the following. This mismatch between the weather forecast and power can be seen in Fig. 3 and Fig. 4. As weather forecasts are valid for a larger area, e.g., 2.8 km, a mismatch between the forecast horizon and the placement of a park causes uncertainty. Furthermore, the uncertainty increases with an increasing forecast horizon of the weather prediction [5].

The solar dataset consists of 21 parks while the wind dataset consists of 45 parks. Both datasets include day-ahead weather forecasts, between 24 and 48h into the future, of the European center for medium-range weather forecast model [26]. The data also includes the corresponding normalized historical power measurements. The solar dataset has a three-hourly resolution for two years and two months. We linearly interpolate the data to have a resolution of one hour to increase the number of samples, especially for the inductive TL problem. The wind dataset has data from two years with an hourly resolution. The data is linearly interpolated to have a 15-minutes resolution. Respectively, the PV dataset has 24 timestamps per day and 96 for the wind dataset. Days not fulfilling these criteria are neglected. In both cases, the first year is considered for training while the remaining data is used for testing. Weather features are standardized based on the training data.

This process results in about 8200 samples per park for training and 9400 for testing in the solar dataset. The wind dataset has about 28400 training and 27400 test samples. The solar dataset contains 38 features, such as sun position and solar radiation. The wind dataset contains seven features, such as wind speed and wind direction. To test the algorithm for inductive TL and zero-shot learning, we split the parks through five-fold cross-validation so that each park is a target task once while having different source tasks.

One disadvantage of these two datasets is that they do not contain any meta-information. Typically, the meta-information in zero-shot learning is used as a proxy to find a similar source task. However, this information is also not available in other open datasets for renewable power forecasts and similarity can also be derived through the input space, see Sec. 4.4. One advantage of the datasets is that they are distributed throughout Germany and even Europe. This distribution makes the MTL, inductive TL, and zero-shot learning more realistic as one cannot assume that parks are always nearby in practice.

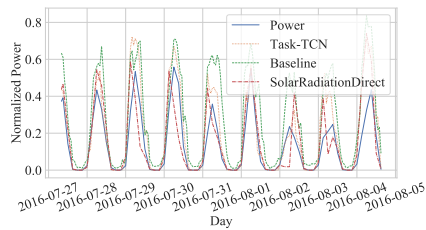
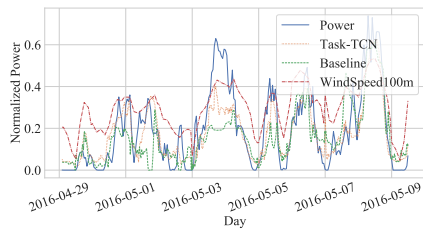

**Fig. 3.** Examplarly forecasts of a solar park with 90 days training data.

**Fig. 4.** Examplarly forecasts of a wind park with 90 days training data.

## 4.2   Evaluation Measures

In all experiments, we are considering the normalized root-mean-squared error (nRMSE). A significant difference between a (reference) model and the baseline is tested on the nRMSEs through the Wilcoxon test (with $\alpha = 0.05$). To calculate improvements against the baseline, we consider the mean skill of all $m$ parks by

$$\text{skill}_m = 1 - \frac{\text{nRMSE}_{\text{reference}_m}}{\text{nRMSE}_{\text{baseline}_m}} \text{ and skill} = \frac{1}{M} \sum_{m=1}^{m=M} \text{skill}_m, \qquad (3)$$

where values larger than zero indicate an improvement upon the baseline.

## 4.3   MTL Experiment

In this section, we conduct an experiment to answer the research question:

*Question 1.* Can a MTL CNN architecture improve the forecast error for wind and PV parks compared to a similar MLP architecture?

**Findings:** The proposed task-TCN improves the baseline up to 18% for the solar dataset and 13% for the wind dataset.

**Design of Experiment:** Ten percent of the training data is used as a validation dataset for hyperparameter optimization in the experiment. Details on the grid search for hyperparameter optimization are available in the supplementary material. Due to the five-fold cross-validation, each park is trained four times, see Sec. 4.1. We calculate the mean nRMSE of a park for these runs.

In total, we train six models and evaluate them through the nRMSE and the skill. The first two are the task embedding MLP and the Bayesian variant of the embedding layer, where the latter is considered the baseline. For the TCN we provide the encoded task ID from the embedding once only to the *first* residual block and in the other variant to *all* except the last layer. We exclude the last layer as we assume that sufficient information from the embedding is available at this point in the network. Afterward, only a transformation to the power of a specific task is required. For each of those two models, we consider a Bayesian and a non-Bayesian variant.

**Detailed Findings:** All variants of the TCN lead to a significant improvement compared to the Bayesian MLP as the baseline, see Table 1. The MLP with a non-Bayesian embedding layer has significantly worse results than the baseline for both datasets. In the case of the PV dataset, the TCN models have mean improvements between nine and eighteen percent. The best TCN model achieves a mean improvement of 25 percent for the wind dataset.

Interestingly, for both datasets, the Bayesian TCNs perform substantially worse than their non-Bayesian variants. The reason probably is that Bayesian models are sensitive to selecting the prior, which was not optimized through grid-search. However, the Bayesian MLP forecast error is significantly better than the non-Bayesian approach. This result is promising as it suggests that it is a strong baseline for the following experiments. The results of the TCN models are promising as well, as they improve upon the baseline. Finally, it is essential to consider that probably due to the larger correlation between PV parks in this dataset, as described in [3], it is beneficial for the TCN to share information in *all* residual blocks. In contrast, the wind dataset tasks are less correlated compared to the PV dataset [3]. Respectively, it is sufficient for the network to only use the first residual block's embedding information.

**Table 1.** Evaluation results for the wind and PV dataset for the mean nRMSE from all parks. The asterisk symbol marks significantly different nRMSE values of reference (Ref.) models than the baseline (BS). The significance is tested through the Wilcoxon signed-rank test with $\alpha = 0.05$. In case that the skill is larger than zero, this indicates a significant improvement upon the baseline.

| | | DataType | PV | | Wind | |
|---|---|---|---|---|---|---|
| | | | Skill | nRMSE | Skill | nRMSE |
| Model | Embedding Type | Embedding Position | | | | |
| MLP | Bayes | First (BS) | 0.000 | 0.087 | 0.000 | 0.184 |
| | Normal | First (Ref.) | -0.087 | 0.095* | -0.239 | 0.229* |
| TCN | Bayes | All (Ref.) | 0.092 | 0.079* | 0.092 | 0.169* |
| | | First (Ref.) | 0.098 | 0.078* | 0.089 | 0.169* |
| | Normal | All (Ref.) | **0.181** | **0.071**\* | 0.254 | 0.137* |
| | | First (Ref.) | 0.174 | 0.072* | **0.258** | **0.136**\* |

### 4.4 Zero-Shot Learning Experiment

In this section, we conduct an experiment to answer the research question:

*Question 2.* Are CNN based MTL architectures capable of providing sufficiently good forecasts in zero-shot learning for renewable power forecasts?

**Findings:** The results let conclude a reasonable forecast error for PV, this does not hold in all cases for the wind dataset due to outliers. Considering that the wind dataset is distributed throughout Europe and the similarity is only on the input space, the results are still promising.

**Design of Experiment:** In this experiment, we use the same model as described in the previous section. Initially, we need to find a suitable source task applicable to an unknown task. As stated earlier in Sec. 4.1, the two datasets do not have meta-information applicable to select an appropriate source task. However, a common approach in TL for time series is to use dynamic time warping (DTW) as a similarity measure in the feature space between the source and the target, see [18]. We assume that a similar park type, e.g., the same rotor diameter, is placed in similar regions with identical weather features. Even though this is not optimal, it is still a reasonable assumption to make. For instance, wind parks near coasts typically have larger rotor diameters than parks placed in a forest. To determine the most similar wind or PV park, we calculate the mean squared difference through DTW between the source and the target. In the case of the wind dataset, we use the wind speed at 100 meter height. For the solar dataset, we utilize direct radiation, as those two are the most relevant features to forecast the expected power [25]. We calculate the similarity based on the first year's training data of the source and target task. Using the entire year is possible, as the input features are themselves forecasts and are extractable for the past. After finding the most prominent candidate through the training data, we forecast on the test dataset. Based on these forecasts, we calculate the nRMSE for each park and the skill, Eq. 3, to measure the improvement.

**Table 2.** Result for zero-shot learning, cf. with Table 1.

| Model | Embedding Type | Embedding Position | PV Skill | nRMSE | Wind Skill | nRMSE |
|-------|----------------|--------------------|----------|-------|------------|-------|
| MLP | Bayes | First (BS) | 0.000 | 0.093 | 0.000 | 0.202 |
|  | Normal | First (Ref.) | -0.055 | 0.098 | -0.159 | 0.226* |
| TCN | Bayes | All (Ref.) | 0.123 | 0.082* | **0.109** | **0.182**\* |
|  |  | First (Ref.) | 0.123 | 0.082* | 0.104 | 0.182* |
|  | Normal | All (Ref.) | **0.129** | **0.081**\* | 0.058 | 0.188 |
|  |  | First (Ref.) | 0.127 | 0.082* | 0.059 | 0.188 |

**Detailed Findings:** Table 2 summarizes the respective results. The baseline has an nRMSE of 0.09 on the PV dataset and TCN models have an error of about 0.08. The non-Bayesian TCN even achieves an improvement of almost 13 percent. For the wind dataset, the improvements range between six and ten percent. The non-Bayesian TCNs and the MLP are not significantly better than the baseline. For all models on the wind dataset, outliers above an nRMSE of 0.5 are present.

These outliers are not surprising, as the wind dataset has parks spread throughout Europe with distinct topographies and weather situations. Furthermore, the source task selection is solely based on DTW, as a proxy for the missing meta-information. We can assume that due to the prior in Bayesian TCNs we have fewer outliers for indistinguishable tasks. The zero-shot learning results for the PV dataset has only a ten percent worse nRMSE compared to the MTL experiment.

### 4.5 Inductive TL Experiment

In this section, we conduct an experiment to answer the research question:

*Question 3.* Are CNN based MTL architectures beneficial during inductive TL compared to a similar MLP approach.

**Findings:** With only 90 days of training data, the nRMSE is similar to the experiment with an entire year of training data. This result shows that the proposed model is extendable to new tasks having a similar error while avoiding catastrophic forgetting with improvements above 25%.

**Design of Experiment:** As seasonal influences affect the forecast error [25], we test how the different seasons as training data result in other target errors. Further, to test the influence of the amount of available training data, we train with data from $7, 14, 30, 60, 90$ and additionally 365 days. To avoid catastrophic forgetting, we only finetune the embedding layer. In this way, the network needs to learn a transformation of the task ID similar to previous source parks and use this encoding. Finally, we tested two methods to initialize the embedding of the new task. The first one uses the *default* initialization strategy of, e.g., pytorch. The other one *copies* the task embedding based on the most similar task through the smallest mean squared error on 10 percent of the training data.

**Detailled Findings:** In the following, we show exemplary results of the spring season for both datasets. Those lead to the best results compared to other seasons for training. When comparing models with another, observations are similar to those presented in the following also for other seasons. The results with a year of training data and other seasons are available in the supplementary material.

The results for the wind dataset are summarized in Table 3. The TCN variants achieve a significant improvement upon the baseline for all different numbers of training data. The non-Bayesian TCN, where we *copy* the most similar embedding vector, has the best mean nRMSE for all training amounts. For most days of training samples, the TCNs, where *all* residual blocks utilize the encoded task ID, have the best nRMSE. Including the encoded task ID solely in the first residual block leads to similar results. In the other cases, including the encoded task ID at the *first* layer leads to the best results. The MLP with a non-Bayesian embedding and the *default* weighting initialization leads to significantly worse results with more than 14 days of training data.

Table 4 summarizes the same results for the solar dataset. A significant improvement is achieved through the non-Bayesian embedding layer for the MLP and *copying* the encoded task ID. All TCN models achieve a significant improvement upon the baseline. In most cases, the TCN which includes the embedding vector at the first residual block and *copies* the embedding vector, has the best or at least a similar result.

Due to these results, we can summarize that the TCN models achieve significant improvements compared to the MLP baselines. This result is not surprising, as, also in other research domains [17, 18], the hierarchical learning structure of CNNs is beneficial for TL in time series. In our cases, the best PV results are when considering the embedding vector solely at the first residual block. It is probably sufficient since the network can bypass information to later layers, without losing information, due to the residual block. However, for the more non-linear problem of wind power forecasts, it is also beneficial to consider the encoded task ID in all residual blocks.

**Table 3.** Mean nRMSE for spring of wind data set. Significant differences of the reference (Ref.) compared to the baseline (BS) is tested through the Wilcoxon signed-rank test with $\alpha = 0.05$ and marked with *.

| Model | Embedding Type | Embedding Position | Embedding Initialization | nRMSE | | | | |
|---|---|---|---|---|---|---|---|---|
| | | | DaysTraining | 7 | 14 | 30 | 60 | 90 |
| MLP | Bayes | First | Copy (Ref.) | 0.233 | 0.229 | 0.213 | 0.206 | 0.186 |
| | | | Default (BS) | 0.234 | 0.229 | 0.213 | 0.205 | 0.186 |
| | Normal | First | Copy (Ref.) | 0.237 | 0.233 | 0.212 | 0.200 | 0.181 |
| | | | Default (Ref.) | 0.255 | 0.247* | 0.229* | 0.217* | 0.200* |
| TCN | Bayes | All | Copy (Ref.) | 0.178* | 0.192* | 0.171* | 0.170* | 0.162* |
| | | | Default (Ref.) | 0.186* | 0.200* | 0.177* | 0.173* | 0.163* |
| | | First | Copy (Ref.) | 0.177* | 0.195* | 0.176* | 0.172* | 0.165* |
| | | | Default (Ref.) | 0.184* | 0.202* | 0.180* | 0.175* | 0.165* |
| | Normal | All | Copy (Ref.) | **0.161*** | **0.173*** | 0.152* | **0.141*** | **0.137*** |
| | | | Default (Ref.) | 0.179* | 0.188* | 0.155* | 0.146* | 0.139* |
| | | First | Copy (Ref.) | 0.163* | 0.181* | **0.148*** | 0.142* | 0.137* |
| | | | Default (Ref.) | 0.187* | 0.199* | 0.162* | 0.152* | 0.142* |

## 5   Conclusion and Future Work

We successfully showed the applicability of the proposed task TCN for wind and PV day-ahead power forecasts. The proposed architecture provides the possibility to solve critical real-world challenges in renewable power forecasts. It provides a framework for MTL, inductive TL, and zero-shot learning and improves the forecast error significantly compared to the Bayesian MLP task embedding as the baseline. By evaluating MTL architectures for zero-shot learning problems, we showed their effectiveness for wind and PV forecasts. Even though we achieve excellent results for the PV dataset, the wind dataset had some outliers due to its dispersion throughout Europe and thus, better accuracies could not be achieved. Considering that the wind dataset is distributed throughout Europe and the similarity was only on the input space, the results are still promising. We will consider this problem by taking additional meta-information into account to select an appropriate source task in future work.

**Acknowledgments** This work results from the project TRANSFER (01IS20020B) funded by BMBF (German Federal Ministry of Education and Research).

**Table 4.** Mean nRMSE for spring of pv data set. Significant difference of the reference (Ref.) compared to the baseline (BS) is tested through the Wilcoxon signed-rank test with $\alpha = 0.05$ and marked with *.

| Model | Embedding Type | Embedding Position | Embedding Initialization | nRMSE | | | | |
|---|---|---|---|---|---|---|---|---|
| | | | DaysTraining | 7 | 14 | 30 | 60 | 90 |
| MLP | Bayes | First | Copy (Ref.) | 0.305 | 0.204 | 0.167 | 0.110 | 0.116 |
| | | | Default (BS) | 0.305 | 0.204 | 0.167 | 0.110 | 0.116 |
| | Normal | First | Copy (Ref.) | 0.170* | 0.192 | 0.096* | 0.086* | 0.089* |
| | | | Default (Ref.) | 0.263 | 0.199 | 0.132 | 0.093* | 0.097* |
| TCN | Bayes | All | Copy (Ref.) | **0.094*** | 0.100* | 0.097* | 0.083* | 0.084* |
| | | | Default (Ref.) | 0.095* | 0.098* | 0.097* | 0.083* | 0.084* |
| | | First | Copy (Ref.) | 0.106* | 0.104* | 0.091* | 0.084* | 0.084* |
| | | | Default (Ref.) | 0.106* | 0.104* | 0.092* | 0.084* | 0.084* |
| | Normal | All | Copy (Ref.) | 0.096* | 0.083* | 0.078* | 0.076* | **0.077*** |
| | | | Default (Ref.) | 0.106* | 0.105* | 0.117 | 0.079* | 0.078* |
| | | First | Copy (Ref.) | 0.106* | **0.081*** | **0.076*** | **0.076*** | 0.078* |
| | | | Default (Ref.) | 0.117* | 0.090* | 0.099* | 0.078* | 0.080* |

# References

1. D. Gielen, F. Boshell, D. Saygin, et al. The role of renewable energy in the global energy transformation. *Energy Strategy Reviews*, 24:38–50, 2019.

2. R. Schwartz, J. Dodge, N. A. Smith, et al. Green AI. *CoRR*, pages 1–12, 2019. arXiv: 1907.10597.

3. J. Schreiber and B. Sick. Emerging Relation Network and Task Embedding for Multi-Task Regression Problems. In *ICPR*, 2020.

4. S. Vogt, A. Braun, J. Dobschinski, et al. Wind Power Forecasting Based on Deep Neural Networks and Transfer Learning. In *18th Wind Integration Workshop*, page 8, 2019.

5. J. Schreiber, A. Buschin, and B. Sick. Influences in Forecast Errors for Wind and Photovoltaic Power: A Study on Machine Learning Models. In *INFORMATIK 2019*, pages 585–598. Gesellschaft für Informatik e.V., 2019.

6. M. Solas, N. Cepeda, and J. L. Viegas. Convolutional Neural Network for Short-term Wind Power Forecasting. In *Proc. of the ISGT-Europe 2019*, 2019.

7. Y. Ju, J. Li, and G. Sun. Ultra-Short-Term Photovoltaic Power Prediction Based on Self-Attention Mechanism and Multi-Task Learning. *IEEE Access*, 8:44821–44829, 2020.

8. T. Shireen, C. Shao, H. Wang, et al. Iterative multi-task learning for time-series modeling of solar panel PV outputs. *Applied Energy*, 212:654–662, 2018.

9. A. S. Qureshi and A. Khan. Adaptive transfer learning in deep neural networks: Wind power prediction using knowledge transfer from region to region and between different task domains. *Computational Intelligence*, 35(4):1088–1112, 2019.

10. L. Cai, J. Gu, J. Ma, et al. Probabilistic wind power forecasting approach via instance-based transfer learning embedded gradient boosting decision trees. *Energies*, 12(1):159, 2019.

11. H. Zang, L. Cheng, T. Ding, et al. Day-ahead photovoltaic power forecasting approach based on deep convolutional neural networks and meta learning. *Int. JEPE*, 118:105790, 2020.
12. X. Liu, Z. Cao, and Z. Zhang. Short-term predictions of multiple wind turbine power outputs based on deep neural networks with transfer learning. *Energy*, 217:119356, 2021.
13. S. Zhou, L. Zhou, M. Mao, et al. Transfer learning for photovoltaic power forecasting with long short-term memory neural network. In *Proc. of the BigComp 2020*, pages 125–132, 2020.
14. S. Tasnim, A. Rahman, A. M. T. Oo, et al. Wind power prediction in new stations based on knowledge of existing Stations: A cluster based multi source domain adaptation approach. *Knowledge-Based Systems*, 145:15–24, 2018.
15. L. Cao, L. Wang, C. Huang, et al. A Transfer Learning Strategy for Short-term Wind Power Forecasting. In *Chinese Automation Congress*, pages 3070–3075. IEEE, 2018.
16. A. S. Qureshi, A. Khan, A. Zameer, et al. Wind power prediction using deep neural network based meta regression and transfer learning. *Applied Soft Computing*, 58:742–755, 2017.
17. R. Ye and Q. Dai. Implementing transfer learning across different datasets for time series forecasting. *Pattern Recognition*, 109:12, 2021.
18. H. I. Fawaz, G. Forestier, J. Weber, et al. Transfer learning for time series classification. In *2018 IEEE BigData*, pages 1367–1376, 2019.
19. S. J. Pan and Q. Yang. A Survey on Transfer Learning. *TKDE*, 22(10):1345 – 1359, 2009.
20. Z. Fuzhen, Q. Zhiyuan, D. Keyu, et al. A comprehensive survey on transfer learning. *Proc. of the IEEE*, 109(1):43–76, 2021.
21. Y. Zhang and Q. Yang. A Survey on Multi-Task Learning. *CoRR*, pages 1–20, 2017. arXiv: 1707.08114.
22. J. Reis and G. Gonçalves. Hyper-Process Model: A Zero-Shot Learning algorithm for Regression Problems based on Shape Analysis. *Journal of Machine Learning Research*, 1:1–36, oct 2018.
23. S. Ruder. *Neural Transfer Learning for Natural Language Processing.* PhD thesis, National University of Ireland, Galway, 2019.
24. C. Guo and F. Berkhahn. Entity Embeddings of Categorical Variables. *CoRR*, pages 1–9, 2016. arXiv: 1604.06737.
25. J. Schreiber, M. Siefert, K. Winter, et al. Prophesy: Prognoseunsicherheiten von Windenergie und Photovoltaik in zukünftigen Stromversorgungssystemen. *German National Library of Science and Technology*, page 159, 2020.
26. European centre for medium-range weather forecasts. http://www.ecmwf.int/, 2020. [Online; accessed 2021-03-30].
27. C. Blundell, J. Cornebise, K. Kavukcuoglu, et al. Weight Uncertainty in Neural Networks. In *32nd ICML 2015*, volume 37, pages 1613–1622, 2015.
28. S. R. Sain and V. N. Vapnik. *The Nature of Statistical Learning Theory.* Springer-Verlag New York, 2006.
29. L. N. Smith. A Disciplined Approach To Neural Network Hyper-Parameters: Part 1. *CoRR*, pages 1–21, 2016. arXiv: 1803.09820.
30. S. Bai, J. Z. Kolter, and V. Koltun. An Empirical Evaluation of Generic Convolutional and Recurrent Networks for Sequence Modeling. *CoRR*, pages 1–14, 2018. arXiv: 1803.01271.

## A   Intro

The following supplementary material provides additional insights and results for the article *"Task Embedding Temporal Convolution Networks for Transfer Learning Problems in Renewable Power Time-Series Forecast"*.

## B   Examples of the Data

In the following, we provide exemplary plots of the EuropeWindFarm and GermanSolar-Farm datasets. In both datasets, the weather forecasts' uncertainty causes a deviation between the weather feature and the power generation.

We provide a scatter plot between one of the essential features and the historical power measurement for each dataset. We can observe the difference between the weather forecast and the actual generated power in the scatter plots. Similarly, we provide a time series plot.

Generally, PV power forecasts are considered a more straightforward problem as the relation between the solar radiation and the generated power is primarily linear. Nonetheless, features of NWP influence each other non-linearly, making it still a challenging problem. In contrast to the solar dataset, the non-linearity between the most crucial feature - wind speed - and the generated power is more non-linear, making it an even more challenging problem.

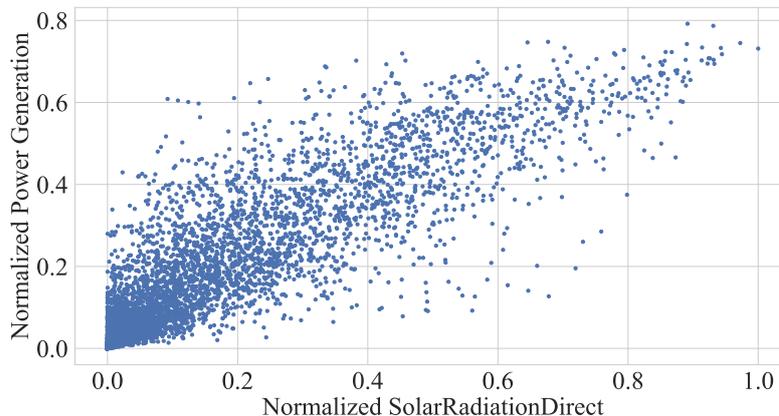

**Fig. 5.** Exemplary scatter plot between the most important feature - radiation - and the generated power for the GermanSolarFarm dataset.

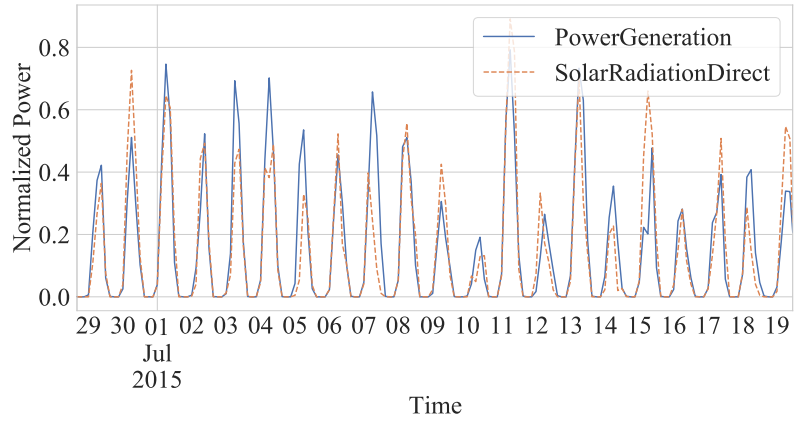

**Fig. 6.** Exemplary time seris plot between the most important feature - radiation - and the generated power for the GermanSolarFarm dataset.

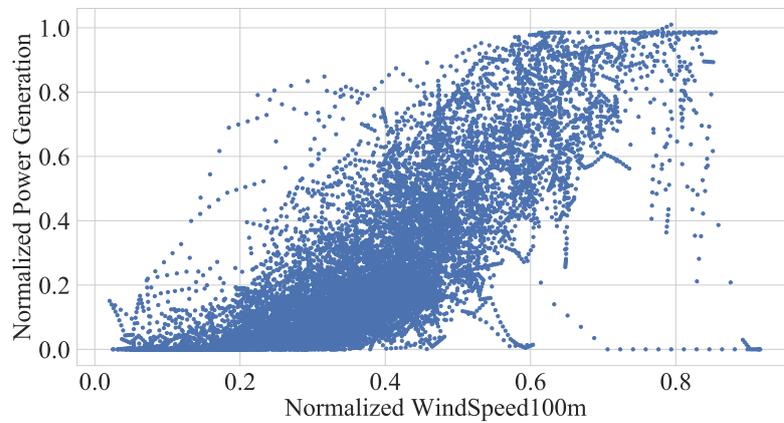

**Fig. 7.** Exemplary scatter plot between the most important feature - wind speed - and the generated power for the EuropeWindFarm dataset.

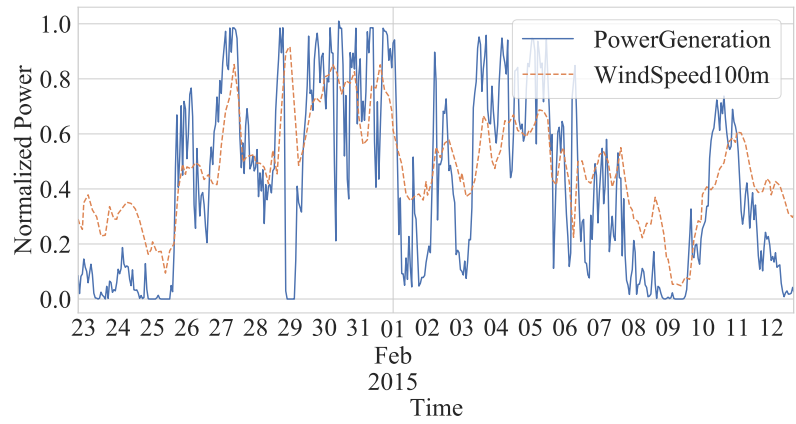

**Fig. 8.** Exemplary time seris plot between the most important feature - wind speed - and the generated power for the EuropeWindFarm dataset.

## C  Method

The following section provides additional material for Sec. 3 - Proposed Method.

### C.1  Task-Embedding for MLPs

Fig. 9 depicts the task embedding MLP. Before training, each task gets a unique task ID assigned. In our case, this is an increasing number as we are interested in extending the MTL to new tasks later on. E.g., when considering 100 PV parks, we have task IDs from one to hundred. This task ID is the input to the embedding layer. For a particular task ID, this embedding space's output is concatenated with other continuous inputs such as those from the NWP. In this way, it is extendable to any network architecture, as the output of the embedding layer can be used as additional continuous input to the network structure. E.g., the network avoids the necessity of task-specific layers of an HPS architecture. However, additional assumptions need to be made; see description of task-TCN in Sec. 3.2.

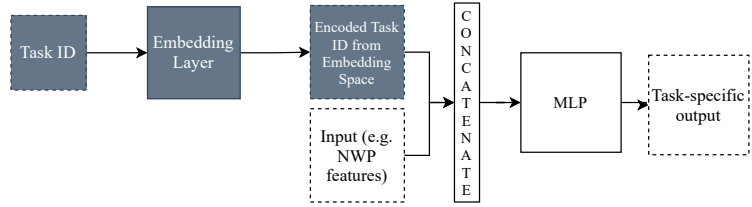

**Fig. 9.** Task embedding for MLP to create task-specific predictions based on a hard parameter sharing architecture without separate layers. By encoding a task ID, for each task, through an embedding layer, the MLP learns task-specific forecasts while utilizing the data from all tasks to improve predictions.

### C.2  Bayesian Task-Embeddings

In our particular approach of the Bayesian task embedding, see Sec. 3.2, we utilize a standard normal distribution as variational distribution. Respectively, each embedding vector requires a mean $\boldsymbol{\mu}_{m\beta}$ and the standard deviation $\boldsymbol{\sigma}_{m\beta}$ for the task $m$. By applying the local reparametrization trick [27] we sample from the variational distribution to obtain:

$$\mathbf{w}_{m_i\beta} = \boldsymbol{\mu}_{m_i\beta} + \boldsymbol{\sigma}_{m_i\beta}\boldsymbol{\epsilon}.$$

Afterward, each sample is concatenated with its respective numerical weather prediction $\mathbf{x}_i^m$. By adding the Kullback-Leibler Divergence (KLD), between the standard normal and the variational distribution, to the mean squared error (MSE) loss $\mathcal{L}$, we obtain the following update rules:

$$\Delta\boldsymbol{\mu}_{m_i\beta} = \frac{\partial L}{\partial \mathbf{w}_{m_i\beta}} + \lambda\boldsymbol{\mu}_{m_i\beta}$$

$$\Delta \boldsymbol{\delta}_{m_i\beta} = \frac{\partial L}{\partial \mathbf{w}_{m_i\beta}} \epsilon + \lambda \left( \boldsymbol{\delta}_{m_i\beta} - \frac{1}{\boldsymbol{\delta}_{m_i\beta}} \right)$$

# D   Experiment

## D.1   Design of Experiment of Sec. 4.3 - MTL Experiment

Based on the description on Sec. 4.3, we now detail the model's hyperparameter optimization via grid search. The layers' initialization strategy is the default one of the utilized libraries (*pytorch*[1], *fastai*[2], and *blitz-bayesian-pytorch*[3]). For all models, we conducted a grid search on the validation dataset. We initially transform the model's number of features in a higher dimension through a factor $k \in \{1, 5, 10, 20\}$. Note that initially transforming the input features in a higher-dimensional space typically improves the performance [28]. Afterward, the features are reduced by 50 percent in each layer to a minimum of 11 before the final output. The final two layers of all networks have sizes 5 and 1. The learning rate of the *adam* optimizer is $10^{-4}$ for all experiments. To accelerate the training, we initially train for 25 epochs with a one-cycle learning rate scheduler [29] with a maximum learning rate of $10^{-3}$ and cosine annealing. During each training, we train for 50 or 100 epochs with a batch size of 1024, 2048, or 4096 for the MLP and 32, 64, or 128 for TCN models as hyperparameters. The initial features of all models are fed through a batch norm. This batch norm assures that input features are within the same range for all parks. Other parameters of the TCN are equal to [30] with a kernel size of three. The MLP has the following components in each layer: batch norm, linear layer, ReLU activation. In the Bayesian variants we weight the KLD by one of $\{10^{-1}, 10^{-3}, 10^{-6}\}$.

## D.2   Design of Experiment of Sec. 4.5 - Inductive TL Experiment

For seven days of training, we used 20 and otherwise 10 percent of the training data to select the number of epochs $(1, 2, 5, 10, 20)$ and the weight decay $(0, 0.25, 0.5)$. Note that we used the entire first year for standardization of the target data. This approach is generally practical as the input features are forecasts for each location with a long history.

---

[1] https://pytorch.org/docs/stable/index.html, accessed 2020-04-29
[2] https://docs.fast.ai/, accessed 2020-04-29
[3] https://pypi.org/project/blitz-bayesian-pytorch/, accessed 2020-04-29

# E    Exploratory Analysis of Task-TCN

The task-TCN initially transforms the discrete task ID through an embedding layer. Ideally, in this **embedding space**, similar tasks should be close to one another. However, in contrast to the task embedding for MLPs, an additional transformation occurs, which results in a **transformed embedding space**. In particular, the embedding space is transformed through a 1D convolution. The results of this transformation are added to the output of the residual block. By adding the convolution results, we essentially add a bias to each channel, depending on the relationship between tasks.

For instance consider, the output of the first residual block has 100 channels of 24 timesteps. For a single sample, the dimension will be $1 \times 100 \times 24$. Also, consider each task has an embedding vector of dimension $1 \times 10$. The embedding tensor will be then $1 \times 10 \times 24$. After applying the transformation, through the 1D convolution, the transformed embedding tensor will also be of dimension $1 \times 100 \times 24$. In the following, we exemplarily visually inspect features of those spaces and compare them with each other. For those visualizations, we use the task-TCN with a non-Bayesian embedding layer which only includes the encoded task ID in the first residual (TCN) block.

### E.1   Visualization of Embedding Space through t-SNE

In the following, we visualize the embedding space and the transformed embedding space through t-distributed stochastic neighbor embedding (t-SNE) in Fig. 10 and 11. Therefore, we create once the embedding for each task. In the example here, this results in a vector of dimension $36 \times 12$. The first dimension is the number of tasks and the second dimension is the resulting dimension of the embedding layer. In the transformed embedding space, we have a dimension of $36 \times 140$, where 140 is now the number of channels in the first layer of the task-TCN.

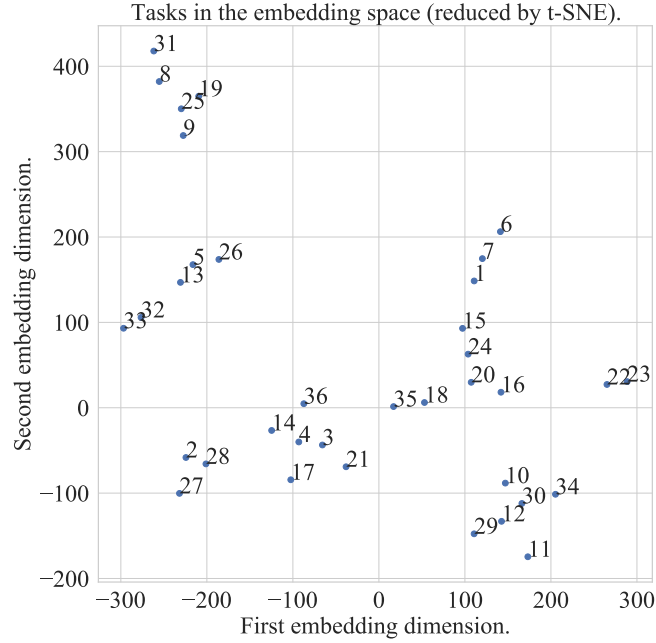

**Fig. 10.** Visualization of the embedding space through t-SNE. The encoded task ID, from the embedding space is calculated for each task and concatenated. The dimensions are reduced through t-SNE and visualized in the scatter plot. Each number corresponds to the respective task ID.

Due to the transformation through t-SNE, we cannot directly compare the embedding and the transformed embedding space. However, we can visually verify if specific tasks are close to one another in both plots. For instance the tasks 29, 12, 30, 10, and 34 are always close to one another in both spaces, c.f. Fig. 10 and 11, even for multiple runs of t-SNE. Similarly, 26, 5, 13, 32, and 33. A smaller distance is also present in various other clusters. However, some cannot be clustered easily. For instance task 11 is close to 29, 12, 30, 34, 10 in the embedding space. In the transformed embedding space, it is closest to 22, 23, 24, 20 and 16. This potentially indicates that task 11 is less similar to other tasks.

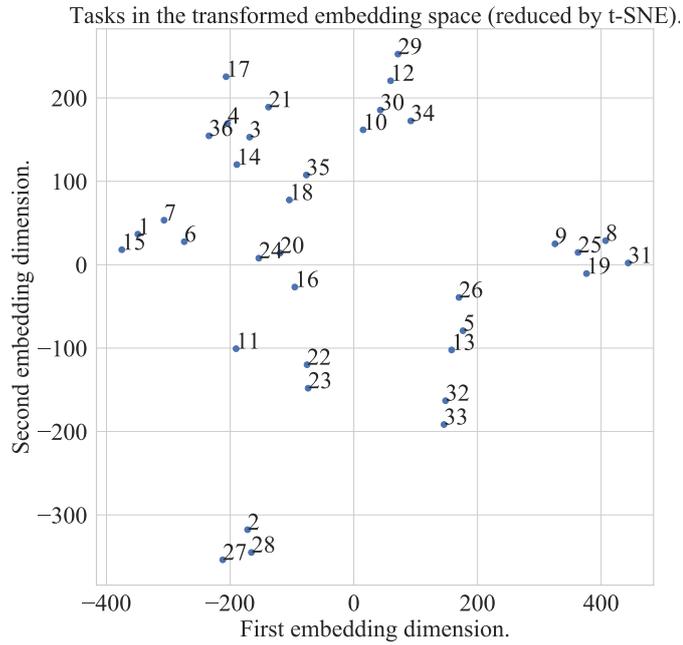

**Fig. 11.** Visualization of the transformed embedding space through t-SNE. The encoded task ID, from the embedding space is calculated for each task and channel. The dimensions are reduced through t-SNE and visualized in the scatter plot. Each number corresponds to the respective task ID.

## E.2   Correlation between Embedding Space and Transformed Embedding Space

To verify if the similarity between tasks in the embedding space is similar to the transformed embedding space, we utilize the Euclidean distance. In particular, we calculate the Euclidean distances for each task with other tasks in the embedding space and the transformed embedding space. Fig. 12 depicts the results of this calulation.

The Pearson correlation of above 0.9 indicates a substantial linear dependency that is also visually available in the scatter plot. These results let us assume that the learned similarity from the embedding space is not drastically changed through the transformation.

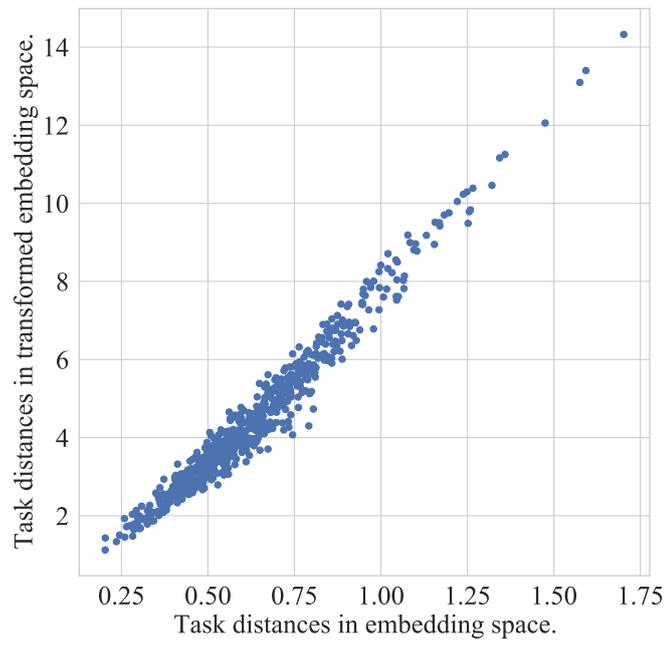

**Fig. 12.** Euclidean distance between task in the embedding space and the transformed embedding space.

### E.3   Distance between Tasks for each Channel

The following exemplary heatmaps visualize the distances between tasks for each channel. Therefore, we calculate the Euclidean distance for each task within a channel. So, we essentially measure the distance of the bias per channel as visualized in Fig. 13, 14, and 15. Interestingly, the similarities or rather dissimilarities between channels differ strongly. For instance, there is a substantial difference of task 30 to all other tasks in channel two. Comparing channel two with channel three, we observe that this pattern is not visible. Instead, there is a substantial dissimilarity between 20 and 21. Other channels such as 9, 10 and 11 have a check-board-like pattern, where we can observe groups of similarities and dissimilarities.

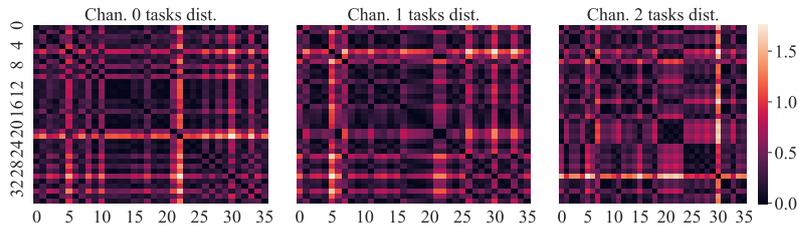

**Fig. 13.** Euclidean distance between task in the transformed embedding space for exemplary channels. Values closer to zero indicate a more substantial similarity.

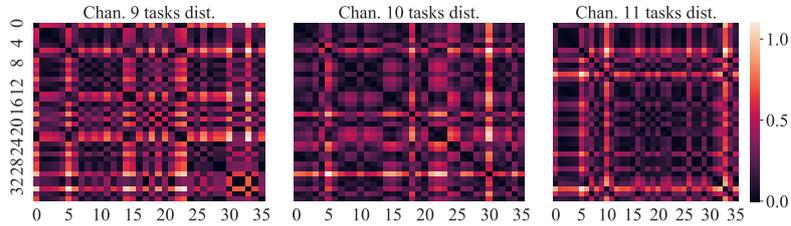

**Fig. 14.** Euclidean distance between task in the transformed embedding space for exemplary channels. Values closer to zero indicate a more substantial similarity.

These findings let us assume that the additional transformation is beneficial in learning complex similarities structures. We can expect that specific channels are responsible for different kinds of weather situations and terrain. Finally, we can conclude that similarities are complex ones that need to be modeled in a high-dimensional space. In this way, the additional transformation is beneficial so that the model can learn diverse features for different weather situations while taking similarities of tasks into account.

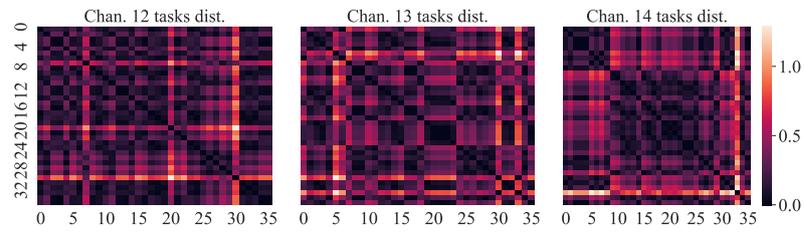

**Fig. 15.** Euclidean distance between task in the transformed embedding space for exemplary channels. Values closer to zero indicate a more substantial similarity.

# F   Exemplary Forecasts of all Models of Sec. 4.5 - Inductive TL Experiment

The following section provides samples plot for both datasets, all six models, and the different initialization strategies for the embedding layer for new parks All models are trained with 90 days from spring.

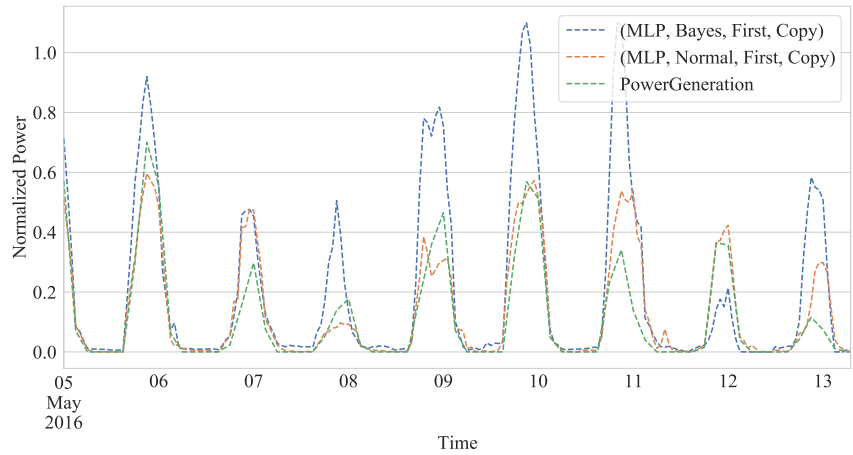

**Fig. 16.** Sample forecast of the task embedding MLP for the GermanSolarFarm dataset. The embedding initialization is copied from the most similar park.

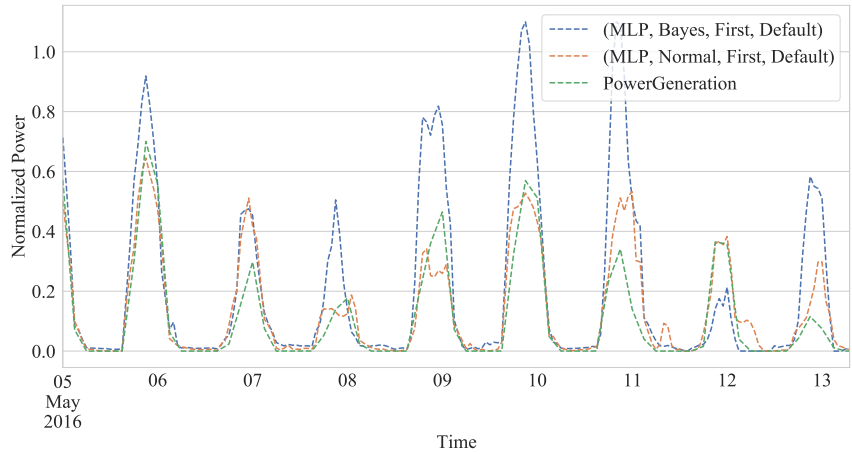

**Fig. 17.** Sample forecast of the task embedding MLP for the GermanSolarFarm dataset. The embedding initialization is the default one.

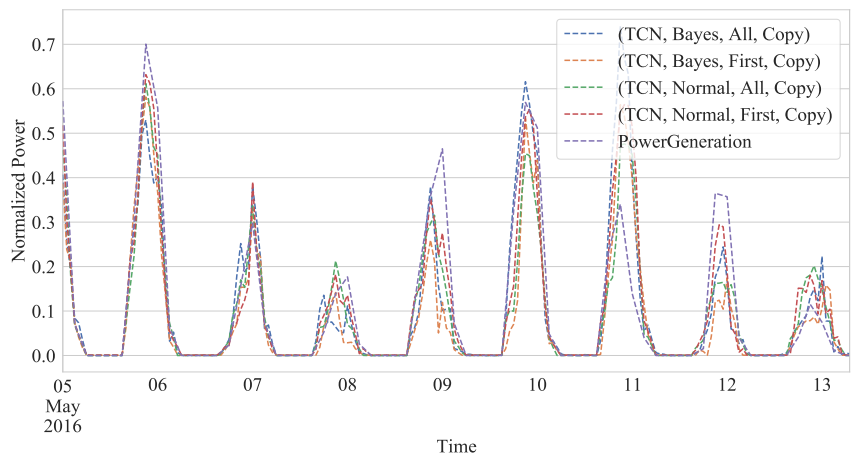

**Fig. 18.** Sample forecast of the task-TCN for the GermanSolarFarm dataset. The embedding initialization is copied from the most similar park.

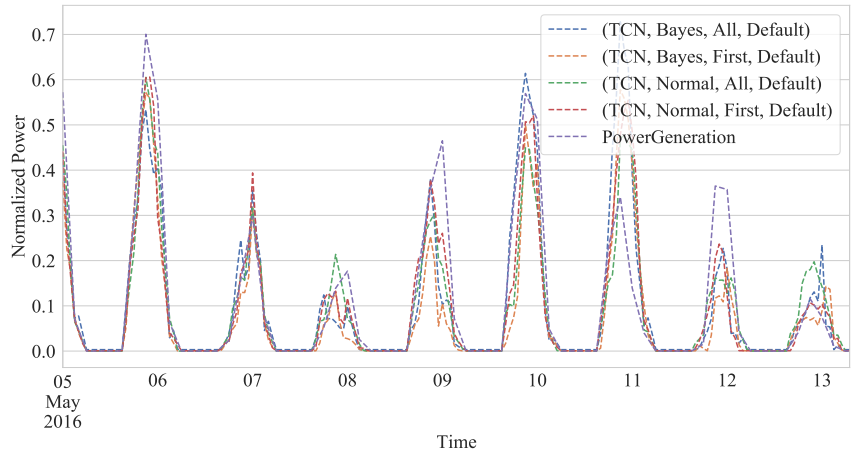

**Fig. 19.** Sample forecast of the task-TCN for the GermanSolarFarm dataset. The embedding initialization is the default one.

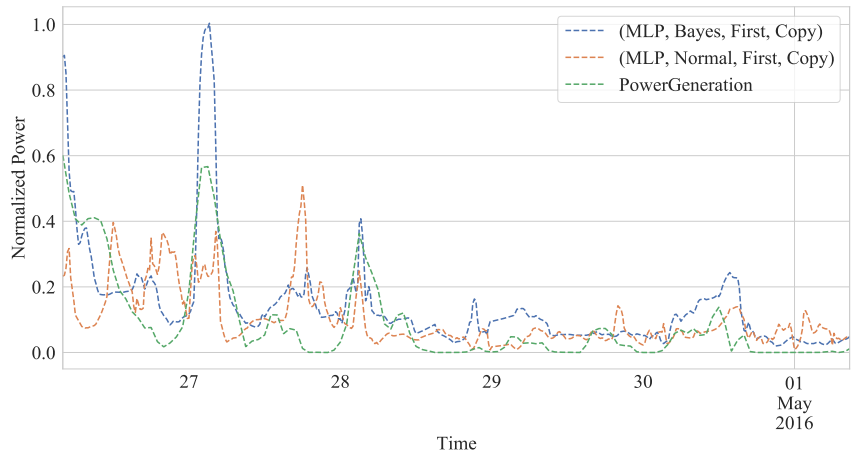

**Fig. 20.** Sample forecast of the task embedding MLP for the EuropeWindFarm dataset. The embedding initialization is copied from the most similar park.

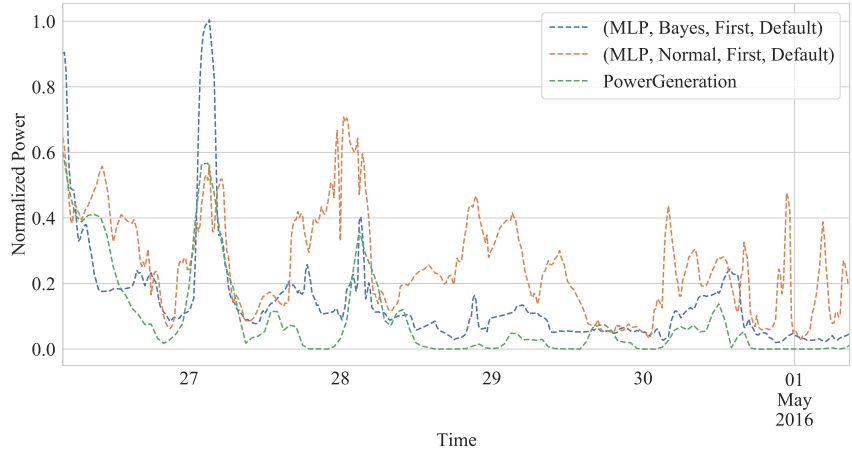

**Fig. 21.** Sample forecast of the task embedding MLP for the EuropeWindFarm dataset. The embedding initialization is the default one.

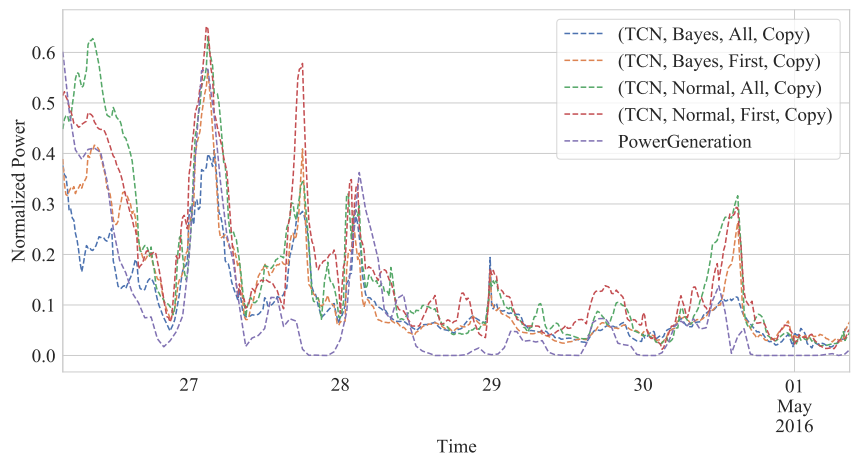

**Fig. 22.** Sample forecast of the task TCN for the EuropeWindFarm dataset. The embedding initialization is copied from the most similar park.

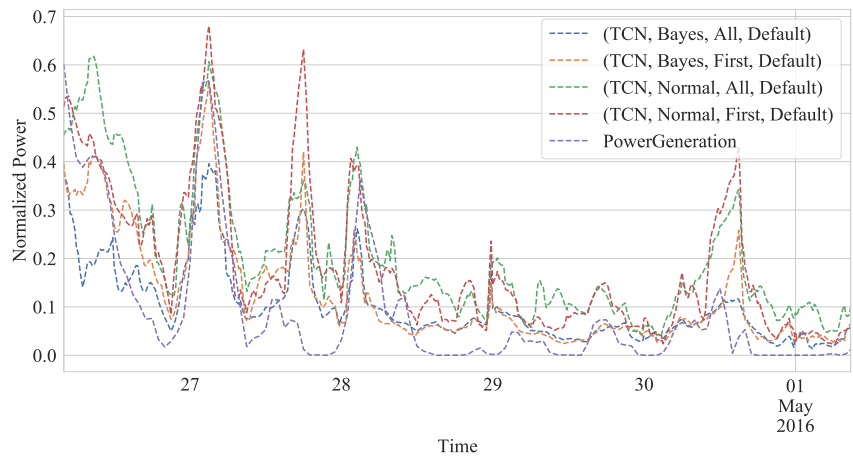

**Fig. 23.** Sample forecast of the task TCN for the EuropeWindFarm dataset. The embedding initialization is the default one.

# G   Detailed Results of Sec. 4.5 - Inductive TL Experiment

## G.1   EuropeWindFarm Results

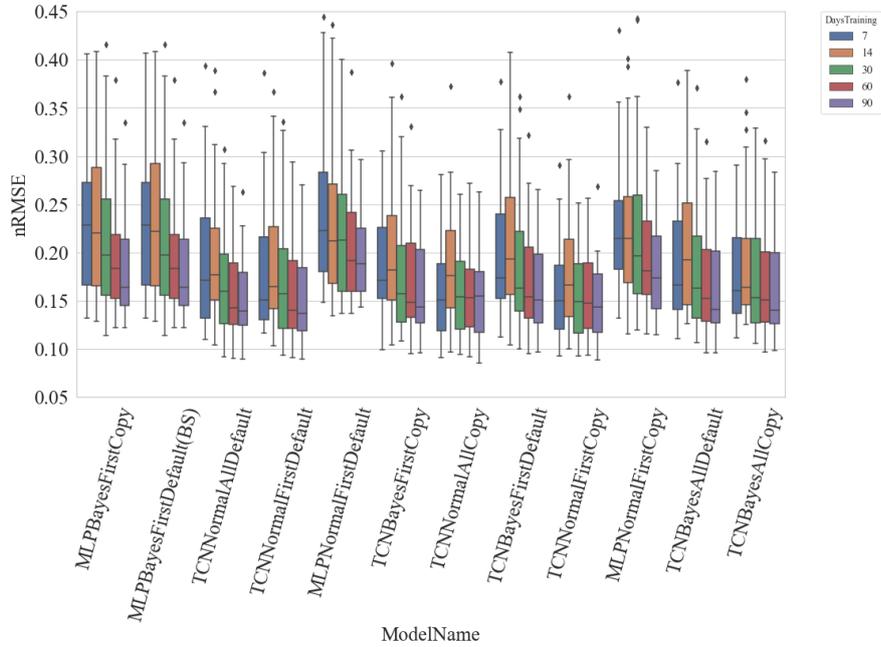

**Fig. 24.** Results for inductive TL experiment for winter of the wind dataset. BS marks the baseline.

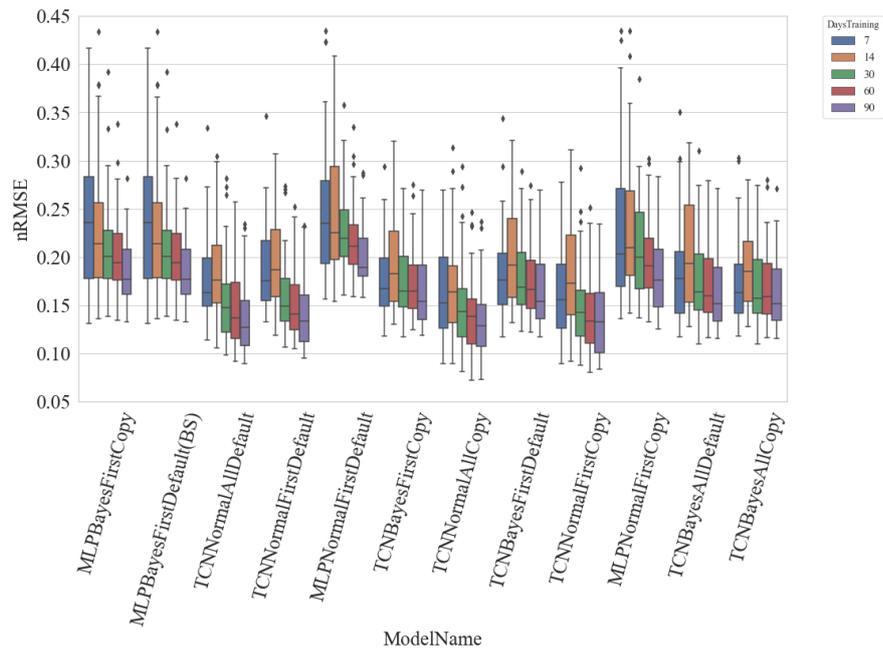

**Fig. 25.** Results for inductive TL experiment for spring of the wind dataset. BS marks the baseline.

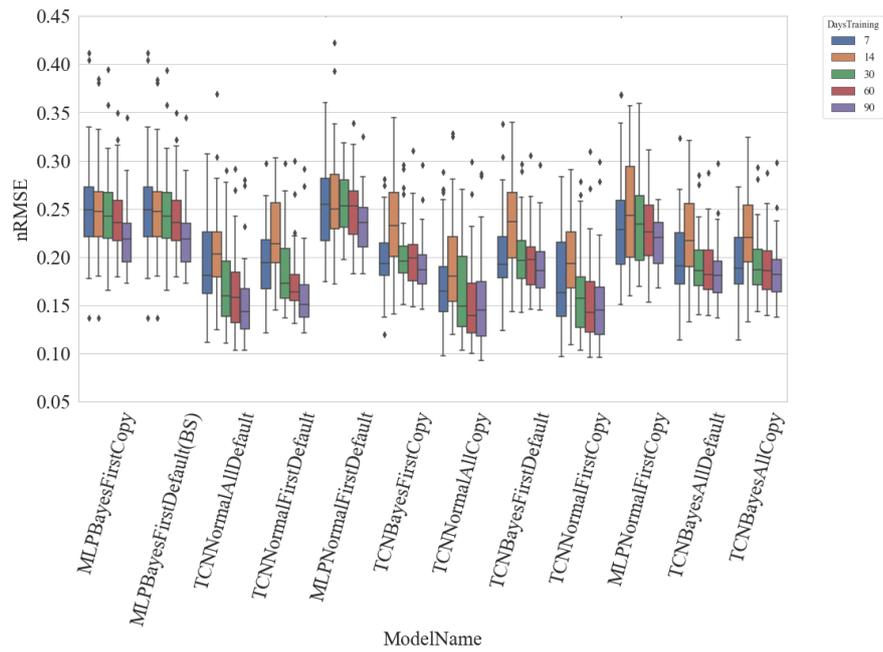

**Fig. 26.** Results for inductive TL experiment for summer of the wind dataset. BS marks the baseline.

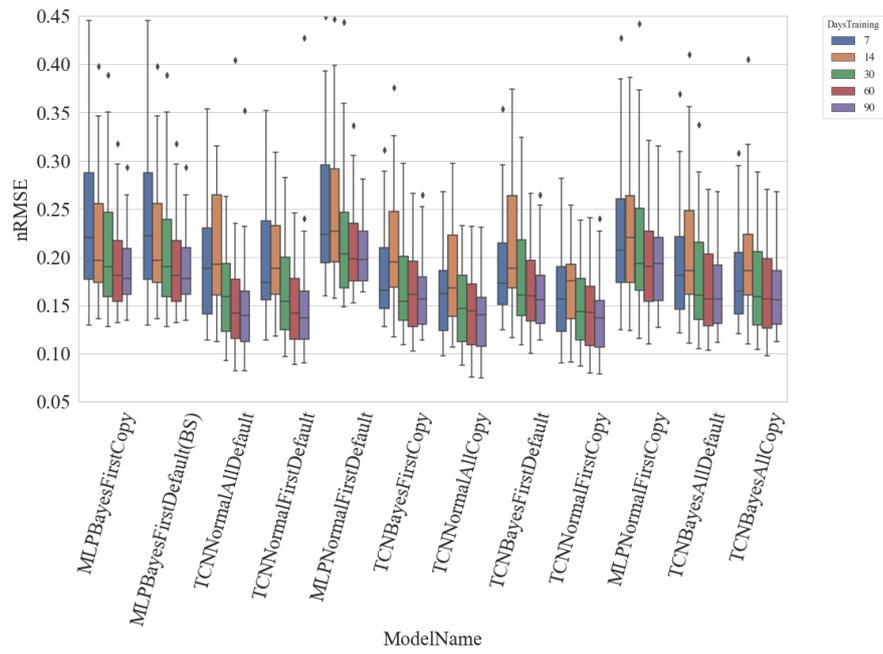

**Fig. 27.** Results for inductive TL experiment for autumn of the wind dataset. BS marks the baseline.

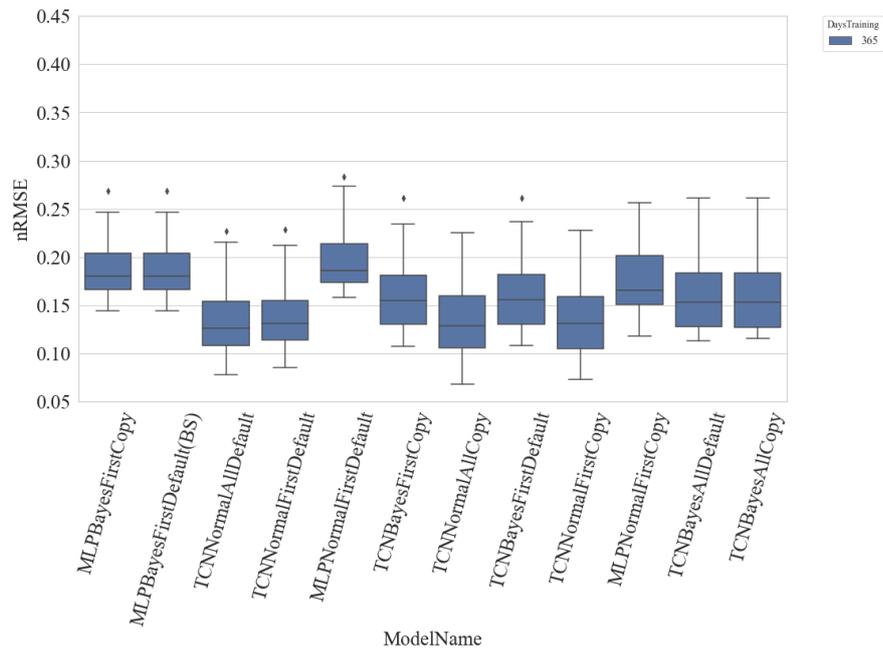

**Fig. 28.** Results for inductive TL experiment for complete year of the wind dataset. BS marks the baseline.

### G.2 GermanSolarFarm Results

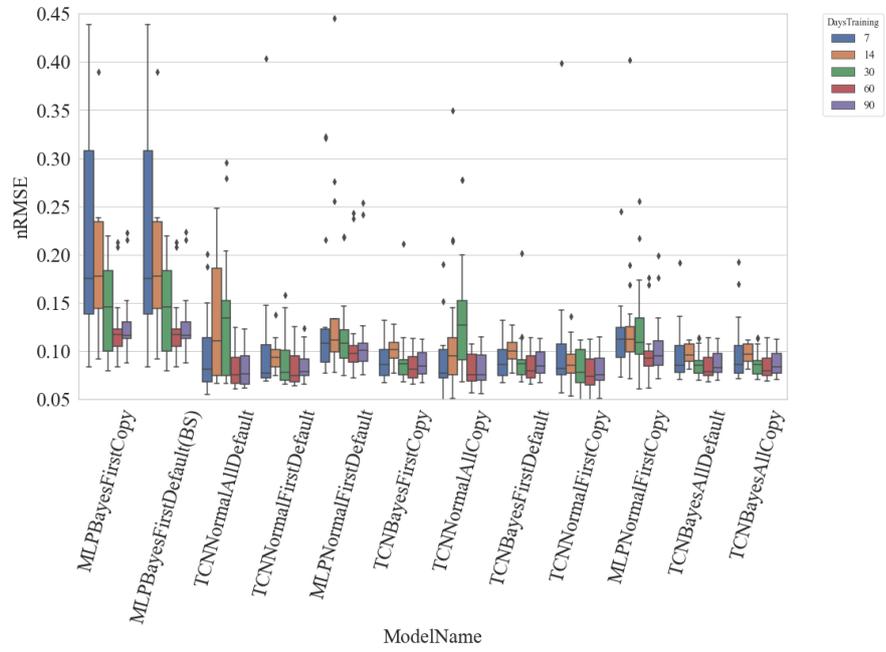

**Fig. 29.** Results for inductive TL experiment for winter of the solar dataset. BS marks the baseline.

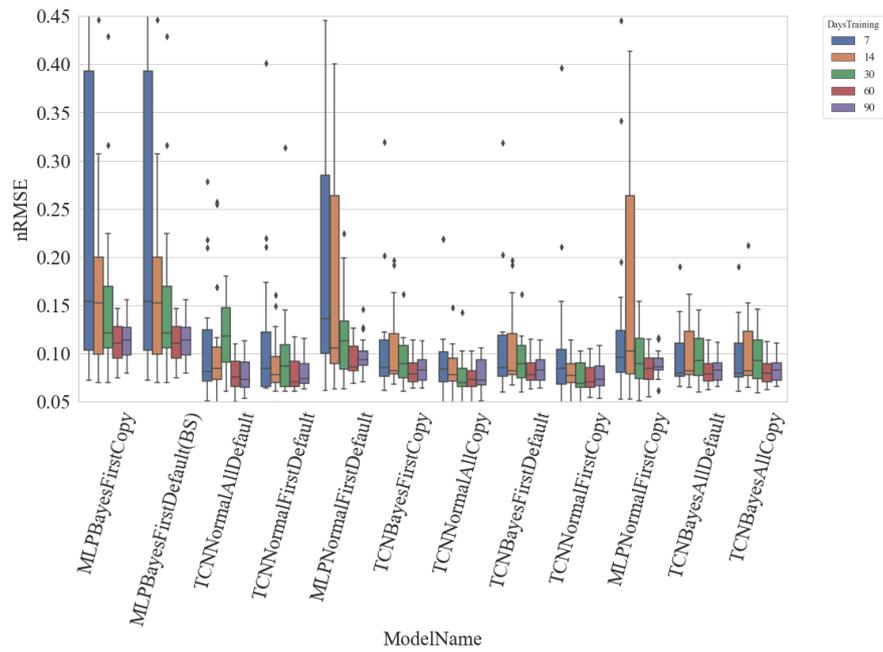

**Fig. 30.** Results for inductive TL experiment for spring of the solar dataset. BS marks the baseline.

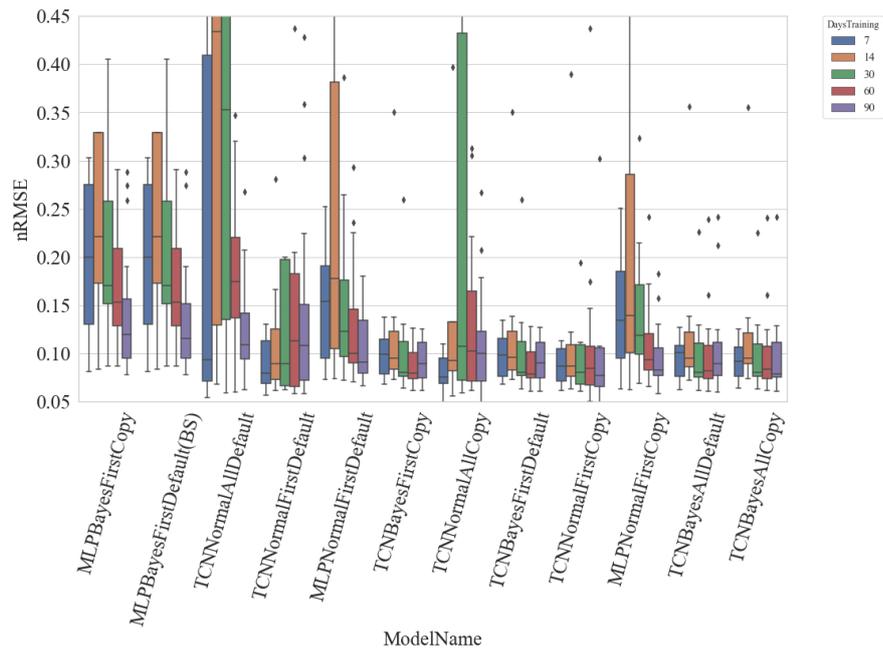

**Fig. 31.** Results for inductive TL experiment for summer of the solar dataset. BS marks the baseline.

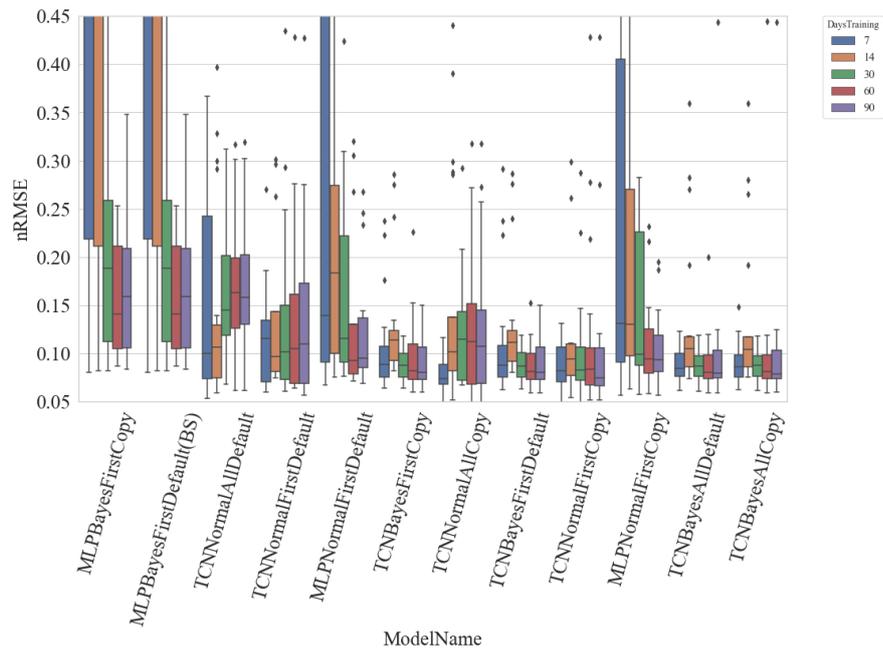

**Fig. 32.** Results for inductive TL experiment for autumn of the solar dataset. BS marks the baseline.

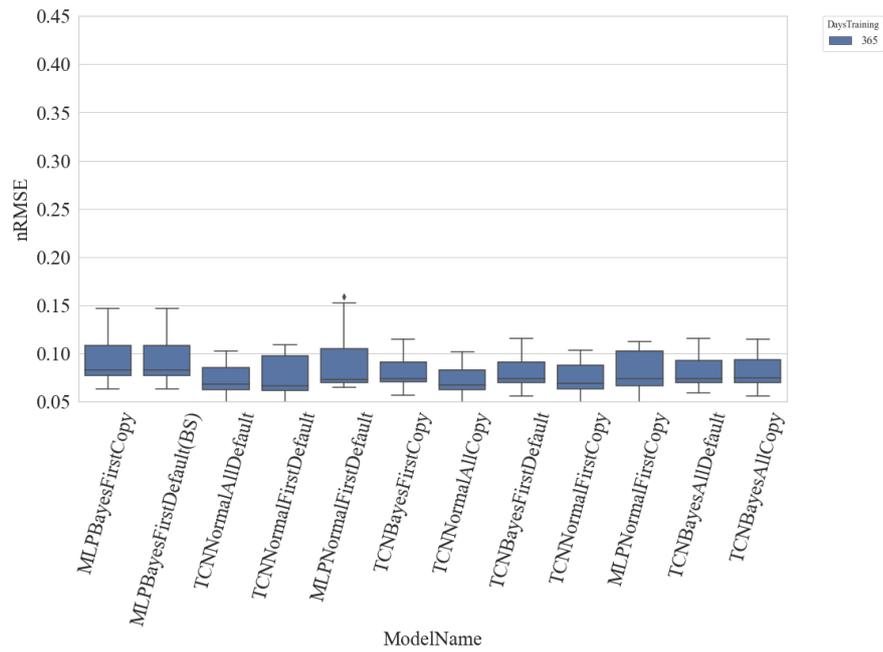

**Fig. 33.** Results for inductive TL experiment for complete year of the solar dataset. BS marks the baseline.